\documentclass[sigconf,authordraft]{acmart}

\AtBeginDocument{%
  \providecommand\BibTeX{{%
    \normalfont B\kern-0.5em{\scshape i\kern-0.25em b}\kern-0.8em\TeX}}}

\usepackage{enumitem}
\usepackage{cuted}
\usepackage{textalpha} 

\begin{document}

\title{LoopAnimate: Loopable Salient Object Animation}

\author{
\normalsize
 Fanyi Wang \footnotemark[1],
 Peng Liu,
 Haotian Hu,
 Dan Meng,
 Jingwen Su,
 JinJin Xu,
 Yanhao Zhang,
 Xiaoming Ren,
 Zhiwang Zhang \footnotemark[1]\\
 OPPO AI Center\\
 11730038@zju.edu.cn\\
}









\begin{strip}
\centering
\includegraphics[width=1.0\textwidth]{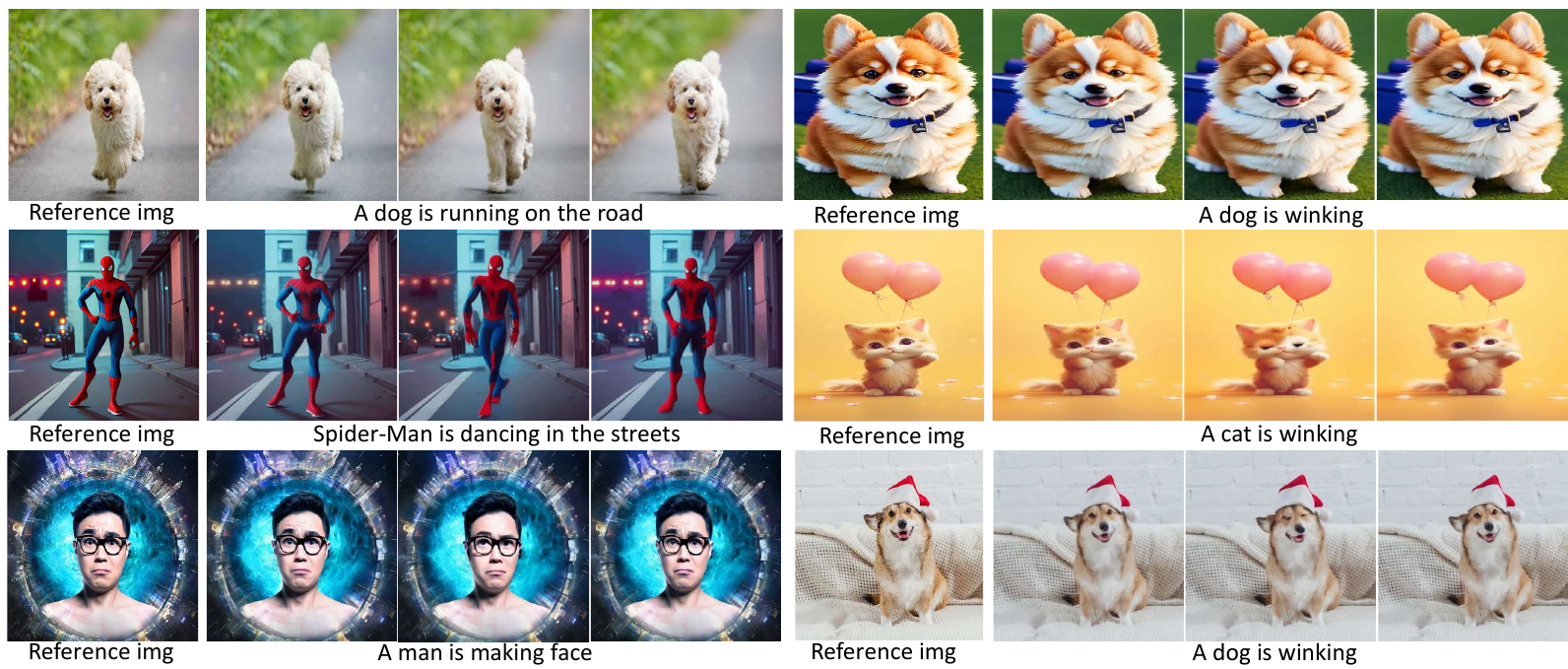} 
\vspace{-2em}
\captionof{figure}{
Generation results of our LoopAnimate with given reference images and text prompts.}
\label{fig:head}
\end{strip}


\begin{abstract}

  
  Research on diffusion model-based video generation has advanced rapidly. However, limitations in object fidelity and generation length hinder its practical applications. 
  Additionally, specific domains like dynamic wallpapers require seamless looping, where the first and last frames of the video match seamlessly. 
  To address these challenges, this paper proposes \textbf{LoopAnimate}, a novel method for generating videos with consistent start and end frames. 
  To enhance object fidelity, we introduce a framework that decouples multi-level image appearance and textual semantic information. Building upon an image-to-image diffusion model, our approach incorporates both pixel-level and feature-level information from the input image, through injecting image appearance and textual semantic embeddings at specific layers of the diffusion model. 
  Existing UNet-based video generation methods use the entire videos as input during training to encode temporal and positional information at once. However, due to limitations in GPU memory, the number of frames is typically restricted to 16. To break this limit, this paper proposes a three-stage training strategy that progressively increases frame numbers and reduces fine-tuning modules. Additionally, we introduce the \textbf{\underline{T}emporal \underline{E}nhanced \underline{M}otion \underline{M}odule} (\textbf{TEMM}) to extend the capacity for encoding temporal and positional information up to 36 frames. 
  The proposed \textbf{LoopAnimate} extends generated video length of UNet-based video generation models to 35 frames while maintaining high-quality video generation. Experiments demonstrate that \textbf{LoopAnimate} achieves state-of-the-art performance in both objective metrics (e.g. fidelity and temporal consistency) and subjective evaluation results.

\end{abstract}






\maketitle

\section{Introduction}
Diffusion model based video generation ~\cite{liu2024sora, guo2023animatediff, chen2023livephoto, blattmann2023stable, wang2023videocomposer, tuneAvideo, magicavatar} is a recent research hotspot filled with challenges. Existing approaches can be categorized into two main architectures: DiT-based ~\cite{peebles2023scalable} and UNet-based ~\cite{ronneberger2015unet}.
DiT-based methods, such as SORA~\cite{liu2024sora} and Latte~\cite{ma2024latte}, utilize Diffusion Transformers (DiT)~\cite{peebles2023scalable} as their backbone. While these methods demonstrate potential for generating longer and higher-quality videos, they also demand significant computational resources and longer inference times.
UNet-based methods, including Animatediff~\cite{guo2023animatediff}, Make Pixel Dance~\cite{zeng2023make}, PIA\cite{zhang2023pia}, and LivePhoto~\cite{chen2023livephoto}, represent the other prevalent approaches. These models typically require inputting the entire video sequence during training and employ motion modules~\cite{guo2023animatediff} for temporal and positional encoding. However, limited by GPU memory constraints, most current works are restricted to generating 16 frames.
Focusing on practical applications, this paper proposes \textbf{LoopAnimate}, a novel image-to-video generation method 
addresses the challenge of generating longer video sequences and enabling seamless transitions between the first and last frames, achieving a total output of 35 frames at once.

To address the challenge of object fidelity in image-to-video generation tasks, we propose the \textit{\underline{M}ultilevel \underline{I}mage representation and \underline{T}extual semantics \underline{D}ecoupling \underline{F}ramework} (\textbf{MITDF}). For long video generation, we introduce a \textit{Three-Stage Training Strategy} to progressively increase frames while reducing fine-tuning modules, along with a \textbf{TEMM} that extends temporal positional encoding capabilities to 36 frames.

Within \textbf{MITDF}, we utilize a pretrained image-to-image diffusion model~\cite{sd-image-variations-diffusers} as the foundation. Input image information is incorporated into the generation process at both pixel-level and feature-level through conditioning and image embedding, respectively. Image representation and textual semantic embedding are injected into different stages of the diffusion model via cross-attention. Experiments demonstrate that excessive image embedding information will limit the motion dynamics of the generated videos, while the middle block contributes most significantly. Consequently, image representation embedding is injected only during the down-sampling process of the diffusion model, and textual semantic embedding is injected solely into the middle and up-sampling blocks. Additionally, we observe that this architecture effectively mitigates the influence of WebVid10M training data ~\cite{Bain21} watermarks during inference.

In the Three-Stage Training Strategy, we first optimize the motion module proposed in Animatediff~\cite{guo2023animatediff} by extending the temporal encoding length to 36 frames. This optimized module is called the \textbf{TEMM}. Then, we carry out a three-stage training process, progressively increasing the number of generated frames in each stage (15, 21 and 35 in the first, second and third stage, respectively) and reducing the number of fine-tuned modules. This enables the model to generate 35-frame videos at a time during the third stage of training. Our contributions can be summarized as follows:
\begin{itemize}[leftmargin=*, topsep=0pt]
\item \textbf{Loopable Video Generation}. We introduce an \textit{\underline{A}symmetric \underline{L}oop \underline{S}ampling \underline{S}trategy} (\textbf{ALSS}) and a specially designed multi-stage conditioning initialization method, enabling the generation of \textit{loopable} videos with salient objects.

\item \textbf{High-Fidelity and Dynamic Motion}. We propose the \textbf{MITDF} to achieve high fidelity in the generated video's object while maintaining excellent motion quality.

\item \textbf{Long Video Generation}. We present a Three-Stage Training Strategy that progressively increases the number of generated frames in each stage while reducing the fine-tuning modules. Combined with the \textbf{TEMM}, which extends the temporal encoding length to 36 frames.
our approach enables the generation of 35-frame videos in a single pass.
\end{itemize}

\section{Related works}


\subsection{Image-to-Video Generation}
Image-to-video generation is to generate a video sequence from a single still image. The goal is to create a realistic and coherent video that depicts motion and temporal changes based on the information present in the input image and text prompt. 

Diffusion models are a class of generative models that can produce realistic images and videos by reversing a stochastic diffusion process. They have been applied to various tasks such as text-to-video ~\cite{khachatryan2023text2videozero, liu2024sora, Align, guo2023animatediff, MakeAVideo, rapheal}, 
image-to-video ~\cite{zhang2023i2vgenxl, shi2024motioni2v, chen2023livephoto, blattmann2023stable, chen2023motion, hu2022make, karras2023dreampose}, 
and video editing ~\cite{ceylan2023pix2video, qiao2024baret, qi2023fatezero, li2024vidtome,ma2023maskint,wang2024stableidentity,bartal2024lumiere}. Some of the recent works that use diffusion models for image-to-video generation are:
\begin{itemize}[leftmargin=*, topsep=0pt]


    

    \item \textbf{I2V-Adapter} ~\cite{guo2024i2vadapter}
    introduces a lightweight plug-and-play module for converting static images to videos while preserving the capabilities of the pretrained Text2Image models. The approach enables interactions between all frames and the noise-free first frame via an attention mechanism, resulting in temporally coherent videos consistent with the first frame.

    \item \textbf{Make Pixels Dance} ~\cite{zeng2023make} addresses the challenge of creating high-dynamic videos, such as those with motion-rich actions and sophisticated visual effects. It can generate videos up to 16 frames long.

    \item \textbf{PIA}~\cite{zhang2023pia} achieves motion controllability by text, and the compatibility with various personalized T2I models without specific tuning. The method introduced the condition module, which takes as inputs the condition frame and inter-frame affinity. It reported a length of up to 16 frames for the output videos.

    \item \textbf{LivePhoto} ~\cite{chen2023livephoto}
    proposes a video generation method that combines three input modalities: reference images, text prompts, and motion dynamics. It also introduce a reweighting of text embeddings to achieve more precise motion control. The method can generate videos with a length of 16 frames.

    \item \textbf{DynamiCrafter} ~\cite{xing2023dynamicrafter}
    can add animation effects to still images in an open domain based on text prompts. It is mainly used in scenarios such as story video generation, loop video generation, and generating frame interpolation. Videos with a length of 16 frames can be generated.

    \item \textbf{Motion-I2V} ~\cite{shi2024motioni2v}
    employs explicit motion field modeling. The model first predicts an explicit motion field, and in the video generation stage, the feature warped by the optical flow field is combined with the original feature to jointly predict the current frame of the video. It can generate videos up to 16 frames long.


    
\end{itemize}

\subsection{Long Video Generation}
For generating long videos, current image-to-video generation methods contain two categories: combining videos generated with multiple inferences and generating as long as possible videos during one inference. The two kinds of methods are complementary and can be used together. The proposed method falls into the second category. Among open-sourced methods, the frame counts of the generated video by one-time inference is highly limited to the memory capacity of the computing devices. Here we list the recent one-time inference methods:

\begin{itemize}[leftmargin=*, topsep=0pt]
    \item \textbf{VideoCrafter} ~\cite{chen2023videocrafter1} is a framework that can generate high-quality videos from text or image inputs, using a diffusion model with a transformer-based ~\cite{transformer} encoder and decoder. It can handle diverse domains such as anime, cartoons, and real-world scenes, and can generate videos up to 256x256 resolution and 32 frames.
    
    \item \textbf{I2VGen-XL} ~\cite{zhang2023i2vgenxl} is a two-stage approach that simultaneously ensures the semantic alignment and the quality of the generated video. The method reported a length of up to 32 frames for the output videos.

    \item \textbf{Stable Video Diffusion} ~\cite{blattmann2023stable} presents a latent video diffusion model for high-resolution, state-of-the-art image-to-video generation. And \textbf{SVD-XT} can generate good quality image-conditioned videos with 25 frames in 2 to 8 steps with 576$\times$1024 resolutions.

    
\end{itemize}

\section{Method}
\begin{figure*}[t]
    \centering
    \includegraphics[width=\linewidth]{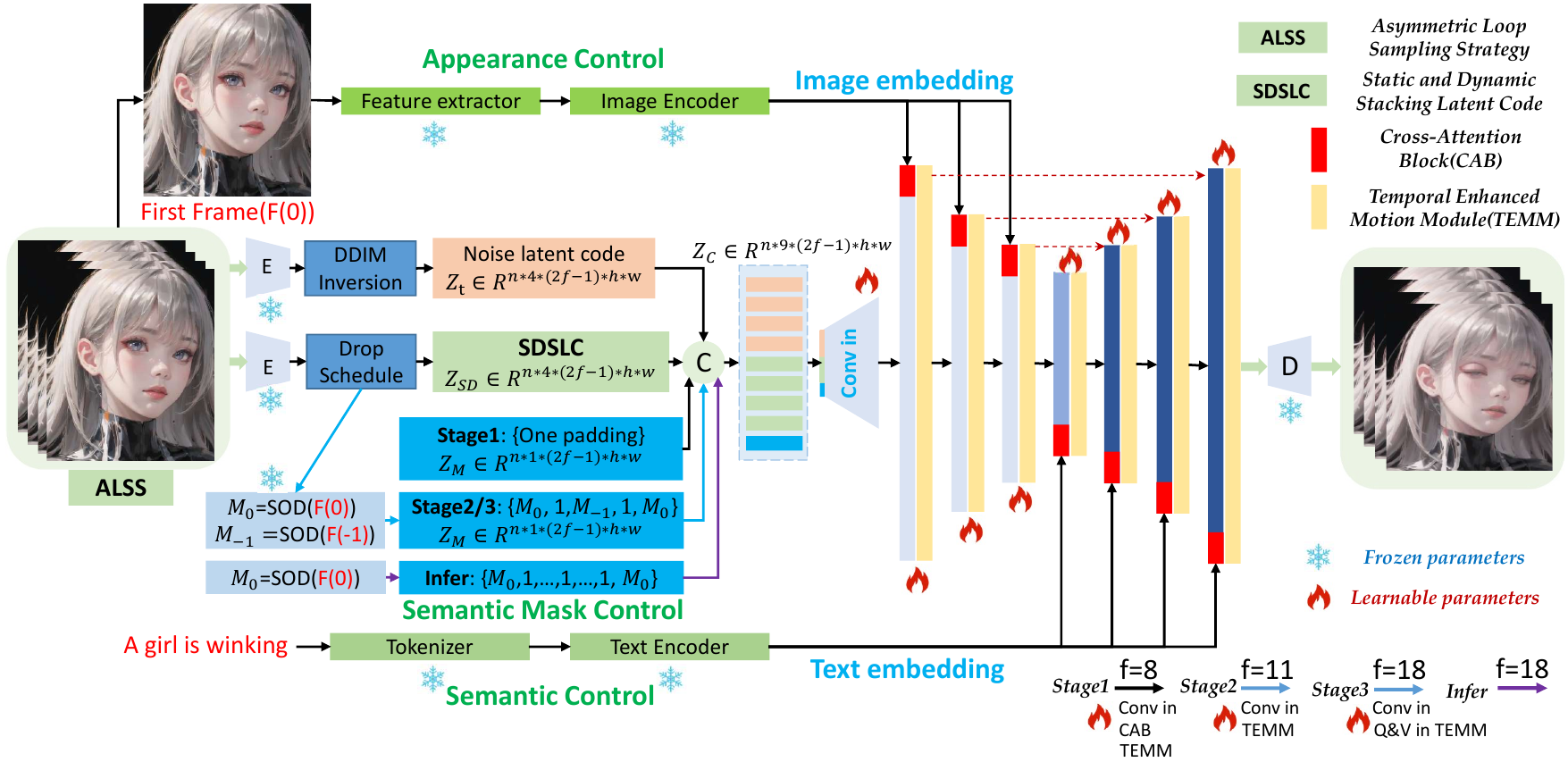}
    \caption{Illustration of LoopAnimate. Arrows for different training stages are with different colors.}
    \label{fig:Fig_flowchat}
\end{figure*}


This paper introduces \textbf{LoopAnimate}, a novel approach capable of generating 35 frames seamlessly looping videos. This is achieved through a combination of techniques, including (1) the Asymmetric Loop Sampling Strategy for training data, (2) a specially designed condition initialization method and (3) Multilevel Image representation and Textual semantics Decoupling Framework(MITDF), and (4) a three-stage training strategy. To ensure consistency between the generated video and the input image, we utilize an image-to-image diffusion model~\cite{sd-image-variations-diffusers}.
And we propose a framework that decouples multi-level image appearance and semantic representation.

\textbf{Multi-level image representation} leverages information from both the pixel-level and feature-level of the input image. As illustrated in Figure.~\ref{fig:Fig_flowchat}, during training, the first and last frames (marked as $F(0)$ and $F(-1)$) of the video are used as one of the conditions to the UNet in the form of \textit{\underline{S}tatic and \underline{D}ynamic \underline{S}tacking \underline{L}atent \underline{C}ode} (\textbf{SDSLC}), enabling pixel-level information injection. Additionally, image embedding is extracted from the input image and cross-attention is performed during the down-sampling process, encoding the image representation information into the latent space and achieving feature-level information injection.

\textbf{Appearance-semantic decoupling} involves injecting the image representation embedding and text semantic embedding at different layers of the denoising network. Inspired by ~\cite{basu2023localizing}, which suggests that different layers of the denoising network in text-to-image tasks have varying influences on style, objects, color, and motion, we conducted ablation experiments in Sec.~\ref{sec:4.2.1}. Similar to text-to-image tasks, we found that the middle block has the most significant impact on motion. We further discovered that excessive insertion of image embedding will restrict the motion dynamics of the generated video. Based on these findings, we finalized the network architecture as shown in Figure.~\ref{fig:Fig_flowchat}. 
Additionally, we increased the temporal position encoding length in the temporal attention module within the motion module, naming it the \textbf{T}emporal \textbf{E}nhanced \textbf{M}otion \textbf{M}odule. We were pleasantly surprised to find that this structure effectively avoids the influence of watermarks present in the WebVid10M training data ~\cite{Bain21} during inference.

In this section, we introduce the (1) fundamentals of diffusion models for image and video generation as a foundation. Subsequently, the proposed method is presented in detail with Figure~\ref{fig:Fig_flowchat}, following the order of the (2) Asymmetric Loop Sampling Strategy for training data, the (3) specially designed condition initialization method and the (4) MITDF, and the (5) three-stage training strategy.

\subsection{Preliminaries}
Diffusion models~\cite{rombach2022highresolution} show promising abilities for both image and video generation. In this work, we opt for a pretrained image-to-image model~\cite{sd-image-variations-diffusers} as the base model, which adapts the denoising procedure in the latent space with lower computations. It initially employs VQ-VAE ~\cite{oord2018neural} as encoder to transform an image ${x}_{0}$ into the latent space: ${{X}_{0}={\varepsilon}({x}_{0})}$. We can sample $X_t$ at any timestep ${t}$ from ${X}_{0}$ directly using a parameterization trick:
\begin{equation}
\label{FL_DROP}
{X}_t = \sqrt{\bar{\alpha_t}} {X}_{0} + \sqrt{1 - \bar{\alpha_t}}\epsilon, ~~ \epsilon\sim \mathcal{N}(\mathbf{0}, \mathbf{I}),
\end{equation}

\noindent where $\bar{\alpha_t}=\prod_{i=1}^t \alpha_i$, $\alpha_t= 1-\beta_t$. and $\beta_t \in (0,1)$ is a predefined noise schedule. The diffusion model uses a neural network $\epsilon_\theta$ to learn to predict the added noise $\epsilon$ by minimizing the mean square error of the predicted noise which writes:
\begin{equation} 
\label{eq:backward}
\min_\theta \mathbb{E}_{X,\epsilon \sim \mathcal{N}(\mathbf{0},\mathbf{I}),t}[\Vert{\epsilon- \epsilon_\theta(X_t,t,{c_t},{c_i})}\Vert_2^2],
\end{equation}
\noindent where ${c_t}$ is semantic embedding encoded from text-prompt, ${c_i}$ is appearance embedding encoded from input image.

\subsection{Asymmetric Loop Sampling Strategy}
To establish a data-level foundation for seamless video generation with matched start and end frames, this paper proposes the Asymmetric Loop Sampling Strategy (ALSS) for processing training video data. ALSS generates training data where the first and last frames are the same, while frames at symmetrical positions in the middle differ, and that's why the number of generated frames is odd. Equation~(\ref{ALS}) illustrates the training data after ALSS processing.
\begin{equation}
\label{ALS}
{L}_{v}=[{F}_{0},{F}_{s},...,{F}_{s(f-1)},{F}_{{s}_{f-1}},{F}_{{s}_{f-2}},...,{F}_{0}],{F}_{-1}={F}_{s(f-1)}
\end{equation}
The number of frames per forward sampling interval is fixed at ${s}$, and the total number of frames for forward sampling is ${f}$. In our experiments, $s=6$. Reverse sampling is performed from ${F}_{-1}$ to ${F}_{0}$, where ${F}_{-1}={F}_{s(f-1)}$ and ${s}_{f-1}$ denotes the number of frames between ${F}_{{s}_{f-1}}$ and ${F}_{0}$. The number of reverse sampling steps is random, subject to the constraint defined in Equation~(\ref{sampling_strategy}).
\begin{equation}
\label{sampling_strategy}
\left\{\begin{aligned}
  &s(f-1)=\sum_{i=1}^{f-1}{s}_{i}-{s}_{i-1}\\
&{s}_{i}-{s}_{i-1} \in [2,4,6,8]
\end{aligned}\right.
\end{equation}

The ALSS, with its stochastic mechanism for reverse sampling steps, fosters better diversity in generated outputs. Compared to post-processing the generated video to loop its motion, the proposed ALSS approach allows for a wider range of motion patterns.

\subsection{Latent Space Conditions Initialization}

\subsubsection{Inverted Noise Latent Code}
Previous experiments and extensive empirical evidence have demonstrated that utilizing noise derived from image degradation as the starting point for noise prediction yields superior results, both in terms of metrics and visual quality, compared to using pure Gaussian noise. Therefore, during the training process, we first employ a VQ-VAE~\cite{oord2018neural} encoder to encode the video into the latent space, as illustrated in Equation~(\ref{video_LATENT}).
\begin{equation}
\label{video_LATENT}
{Z}_{0}={\varepsilon}({F}_{0},...,{F}_{-1},...,{F}_{0})),
\end{equation}
\begin{equation}
\label{video_noise_LATENT}
{Z}_{t} = \sqrt{\bar{\alpha_t}} {Z}_{0} + \sqrt{1 - \bar{\alpha_t}}\epsilon, ~~ \epsilon\sim \mathcal{N}(\mathbf{0}, \mathbf{I}).
\end{equation}

In Equation~(\ref{video_LATENT}), $\varepsilon$ represents the encoder of VQ-VAE, while ${F}_{0}$ and ${F}_{-1}$ denote the first (last) and last (first) frames of the video in forward (reverse) sampling, respectively. ${Z}_{0}$ is the latent code obtained after encoding the video into the latent space. Subsequently, Gaussian noise is added to ${Z}_{0}$ based on Equation~(\ref{video_noise_LATENT}) to acquire the inverted noise latent code ${Z}_{t}$. Notably, ${Z}_{t} \in \mathcal{R}^{n*4*(2f-1)*h*w}$.

\subsubsection{Static and Dynamic Stacking Latent Code}

To enhance motion dynamics while preserving subject fidelity, we hypothesized that the UNet input condition should incorporate noise-free information from the original video during training. Based on this idea, we conducted a series of experiments. Limited results indicated that utilizing only the latent code of the first frame (${F}_{0}$) could achieve subject fidelity to a certain extent.

The most straightforward approach to increase motion dynamics involved incorporating the latent code of the last frame $F_{-1}$. However, considering that the inference process only has access to the initial frame, it is impossible to construct corresponding motion dynamics information between the first and last frames. Therefore, we randomly dropped the last frame information with a fixed probability $r$ during training, as represented mathematically in Equation~(\ref{FL_DROP}). A higher value of $r$ indicates a higher probability of dropping the last frame, which is then filled with zeros. In our experiments, we set $r=0.5$.
\begin{equation}
\label{FL_DROP}
{\mathbb{F}{({F}_{-1},r)}=Drop{({\varepsilon}({F}_{-1}),r)}},
\end{equation}

The \textbf{SDSLC} ${Z}_{SD}$ is defined in Equation~(\ref{SDSL}), where zero padding is applied to all frames except for the first and last frames of the input video.
\begin{equation}
\label{SDSL}
{{{Z}_{SD}} = {[{\varepsilon}({F}_{0}), ZeroPAD,{\varepsilon}(\mathbb{F}{({F}_{-1},r)}), ZeroPAD,{\varepsilon}({F}_{0})]}},
\end{equation}
${{Z}_{SD}} \in \mathcal{R}^{n*4*(2f-1))*h*w}$, ${ZeroPAD}$ are zero padding results, and $\in \mathcal{R}^{n*4*(f-2)*h*w}$. During the inference stage, only the input image information is available. Therefore, the latent code ${Z}_{SD} = {[{\varepsilon}({F}_{0}), ZeroPAD, {\varepsilon}({F}_{0})]}$.


\subsubsection{Fine-Grained Semantic Mask}

In real-world scenarios, the foreground object and background often exhibit significant differences in their motion magnitude and patterns. To effectively capture this distinction during training, we introduce the \textbf{Fine-Grained Semantic Mask} (\textbf{FGSM}) to encode the object's location information at the pixel level. During the training process, we leverage a \textbf{Salient Object Detection} (\textbf{SOD}) algorithm ~\cite{liu2024lightweight} to decouple the foreground and background regions. This decoupling is achieved by applying SOD algorithm ~\cite{liu2024lightweight} to the first and last frames of the video ($F_0$ and $F_{-1}$, respectively). As shown in Equation~(\ref{SOD_mask}), the decoupling process yields binary mask images $M_0$ and $M_{-1}$, which differentiate the foreground from the background.
\begin{equation}
\label{SOD_mask}
\left\{\begin{aligned}
&{M}_{0}={SOD({F}_{0}),M_{0}\in{0,1}}\\
&{M}_{-1}={SOD({F}_{-1}),M_{-1}\in{0,1}}.
\end{aligned}\right.
\end{equation}
In the three-stage training process, the WebVid10M dataset ~\cite{Bain21} used in the first stage covers a broad domain, and videos may not necessarily contain an object. Therefore, ${Z}_{M}$ is uniformly set to $1$ to learn the overall semantic information of the entire image. The second and third stages are trained on cleaned data containing objects, performing \textbf{FGSM} to encode on the corresponding positions of the first and last frames in the input video. As is shown in Equation~(\ref{ZM_init}), where the first and last positions in ${Z}_{M}$ are encoded with ${M}_{0}$. Correspondingly, the ${F}_{-1}$ position in ${Z}_{SD}$ is encoded with ${M}_{-1}$, and all other positions are encoded with $1$.
\begin{equation}
\label{ZM_init}
\left\{\begin{aligned}
&Stage1:{Z}_{M}={[OnePAD]},\\
&Stage2/3:{Z}_{M}={[{M}_{0},OnePAD,{\mathbb{F}{({M}_{-1},r)},OnePAD,{M}_{0}]}}.
\end{aligned}\right.
\end{equation}


During the inference stage, the object's mask $M_0$ is obtained from the input image. We then create a tensor $Z_M$ by replicating one padding to all frames except for the first and last frames ${Z}_{M}={[{M}_{0},OnePAD,{M}_{0}]}$. Finally, $Z_t$, $Z_{SD}$, and $Z_M$ are concatenated along the channel dimension to form the final input condition ${Z}_{C}$ for the UNet, denoted as Equation~(\ref{ZC}), where ${{Z}_{C}} \in \mathcal{R}^{n*9*(2f-1)*h*w}$,
\begin{equation}
\label{ZC}
{Z}_{C} =[{Z}_{t},{Z}_{SD},{Z}_{M}].
\end{equation}

\subsection{Multi-level Image representation and Textual semantics Decoupling Framework}
\label{sec:3.4}
Object fidelity is a crucial aspect of image-to-video generation. To address this challenge, we propose a framework built upon a pretrained image-to-image diffusion model which initially utilizes image embedding as the guidance, and we enable the capability of decoupling multi-level image appearances and textual semantics. This decoupling involves injecting image appearance embedding and text semantic embedding at different layers of the denoising network. Multi-level image representation refers to incorporating pixel-level representation information of the input image during condition initialization, see Equation~(\ref{SDSL}). And extract image embedding from the input image to perform cross-attention to inject the feature-level information into the latent space. Our ablation experiments reveal that, similar to text-to-image tasks~\cite{rombach2022highresolution, pixart-alpha, Imagen, rapheal,qu2023layoutllm,sun2023sgdiff}, text semantics in image-to-video tasks exert the most significant influence in the middle block. Through limited ablation experiments in Sec.~\ref{sec:4.2.1}, we ultimately determined the optimal structure representation to be Equation~(\ref{DOWNSAMPLE_cross}) and Equation~(\ref{UPSAMPLE_cross}).
\begin{equation}
\label{DOWNSAMPLE_cross}
Attn_{D}={softmax}(\frac{{{Q}_{Z}^{D}} \cdot {{K}_{I}^{T}}} {\sqrt{d}}) \cdot {V_I}.
\end{equation}
\begin{equation}
\label{UPSAMPLE_cross}
Attn_{M,U}={softmax}(\frac{{{Q}_{Z}^{M,U}} \cdot {{K}_{T}^{T}}} {\sqrt{d}}) \cdot {V_T}.            
\end{equation}

${Q}_{Z}^{D}$ represents the query extracted from the latent code ${Z}$ of the down-sample layers, while ${K}_{I}^{T}$ and $V_I$ denote the key and value obtained from the image embedding, respectively. Feature-level information from image embedding is injected during downsampling, while textual information from text embedding is incorporated within the middle and upsampling blocks.

$Attn_{D}$ represents the cross-attention in the down-sampling layers, while $Attn_{M,U}$ denotes the cross-attention in both the middle and up-sampling layers. ${Q}_{Z}^{M,U}$ represents the query extracted from the latent code ${Z}$ of the middle, up-sample layers, while ${K}_{T}^{T}$ and $V_T$ denote the key and value obtained from the text embedding, respectively. Further details can be found in Equation~(\ref{qkv_DM}).
\begin{equation}
\label{qkv_DM}
{{Q}_{Z}^{D}= {Z}{W_q}, {K_I} = {c_i}{W_k}, {V_I} = {c_i}{W_v}}.
\end{equation}

Similar to Equation~(\ref{qkv_DM}), as shown in Equation~(\ref{qkv_U}), 
${Q}_{Z}^{M,U}$ represents the query extracted from the latent code ${Z}$ of the middle and up-sample layers, while ${K}_{T}$ and $V_T$ denote the key and value obtained from the text embedding, respectively.
\begin{equation}
\label{qkv_U}
{{Q}_{Z}^{M,U}= {Z}{W_q^{'}}, {K_T} = {c_t}{W_k^{'}}, {V_T} = {c_t}{W_v^{'}}}.
\end{equation}
${{W_q}, {W_k}, {W_v}}$ are the weight matrices of the trainable linear projection layers in down-sample blocks, and ${{W_q^{'}}, {W_k^{'}}, {W_v^{'}}}$ are in middle  and up-sample blocks. ${{c_i},{c_t}}$ are image embedding and text embedding separately.

\subsection{Three-Stage Training Strategy}
\label{sec:3.5}

Existing UNet-based video generation models require the input of an entire video for temporal and positional information encoding. However, limited by hardware memory, most existing video generation works can only generate $16$ frames at a time. To address this issue, we propose a three-stage training strategy which progressively increases the number of generated frames while reducing the fine-tuning modules in each stage. Additionally, we introduce the \textbf{TEMM}, which extends the temporal and positional information encoding length to $36$ frames. 

\textit{Stage1.} To ensure object consistency, we incorporate the latent code information of the original image into $Z_{SD}$ and positional encoding information into $Z_{M}$ during condition initialization. In the first stage, we train the \textit{conv in}, \textit{cross-attention block}, and \textit{TEMM} with $f=8$ (15 frames). \textbf{TEMM} expands the positional embedding length to 36 based on the motion module, leaving the remaining structure of the motion module unchanged. 
When training \textit{conv in}, we reuse the weights of the $4$ channels from the pretrained image-to-image model and randomly initialize another $5$ channels for the inputs of $Z_{SD}$ and $Z_{M}$. All $9$ channels of \textit{conv in} are optimized simultaneously during training. The first stage of training is conducted on the full WebVid10M dataset ~\cite{Bain21} to align the input $Z_C$, image embedding, and text embedding, and to enable \textbf{TEMM} to acquire temporal encoding capabilities.

\textit{Stage2.} When the input image contains a salient object, maintaining object consistency and action continuity in the generated results becomes crucial, increasing the difficulty of this specific type of video generation task. To tackle this challenge, we leverage SOD ~\cite{liu2024lightweight} to determine whether the first frame of the video contains the same object as described in the caption. To identify whether the object exists within a video, we employed optical flow ~\cite{teed2020raft}. This process allows us to curate the Salient dataset, which contains 94,686 high-quality video segments. In the second stage, we train \textit{conv in} and \textbf{TEMM} with $f=11$ ($21$ frames). Since the change of $f$ alters the encoding length of the positional embedding, \textbf{TEMM} needs to be fine-tuned. Additionally, fine-tuning \textit{conv in} increases negligible computational cost but can accelerate convergence.

\textit{Stage3.} The temporal attention block in \textbf{TEMM} involves computation of Q, K, and V, which consumes a significant amount of GPU memory. Our analysis reveals that, K and V are derived from the same feature map through two MLPs. Consequently, the multiplication operation between Q and K can be optimized by focusing on just one of these variables without compromising the outcome.
Therefore, we only optimize the \textit{conv in}, Q and V within the \textbf{TEMM} during the third stage of fine-tuning. This optimization allows us to increase $f$ to 18, resulting in a total of 35 generated frames.

\section{Experiments}
We employ a pretrained image-to-image model~\cite{sd-image-variations-diffusers} as our foundational model. The training process utilizes 8 A100 GPUs, each equipped with 80GB of memory. We set a batch size of 1 per GPU, and resize each frame proportionally before center cropping it to a resolution of $512\times512$.

\textit{Stage 1.} The model is initially trained on the WebVid-10M dataset ~\cite{Bain21} with $f=8$, enabling the generation videos of 15 frames. The learning rate is fixed at $2.e-4$, and the \textit{conv in}, \textit{cross attention}, and \textbf{TEMM} are trained for 200k iterations.

\textit{Stage 2.} The model is fine-tuned on a processed dataset of 94,686 high-quality videos containing salient objects (Salient Dataset). The \textit{conv in} layer and \textbf{TEMM} are fine-tuned with $f=11$, enabling the generation of 21 frame videos. The learning rate is fixed at $6.e-5$, and the fine-tuning process runs for 100k iterations.

\textit{Stage 3.} The model is further fine-tuned on the Salient Dataset, specifically focusing on \textit{conv in} and the Q and V components of the temporal attention module within the \textbf{TEMM}. We increase $f$ to 18, enabling the generation of 35-frame videos. The learning rate is set to $2.e-5$, and the fine-tuning process runs for another 100k iterations. Table~\ref{tab:parameters} shows the number of parameters fine-tuned in each stage. In Stage 3, only the \textit{conv in} layer, TEMM.Q, and TEMM.V are fine-tuned, totaling $303.2M$ parameters.
The number of frames selected for the three-stage training process is flexible. However, an excessively large number of frames in the first stage can prolong the training time. The temporal positional encoding length in \textbf{TEMM} can be extended, as long as it doesn't exceed the GPU memory limit during the third stage fine-tuning. In our tests, using an A100 GPU with 80GB memory to run a single batch without any engineering optimization, the memory limit was reached when the third stage fine-tuning length was around 35 frames.

For evaluation, we constructed a test set of 99 images with salient objects. These images were sourced from personal photo albums, the Internet, and generated using Midjourney ~\cite{midjourney}. As ground truth videos were unavailable, we employed objective metrics inspired by existing methods like LivePhoto ~\cite{chen2023livephoto} and PIA~\cite{zhang2023pia}. Specifically, we utilized CLIP-I for object fidelity, Frame Consistency (FC) for inter-frame coherence, and SSIM ~\cite{wang2004image} for measuring motion magnitude in the generated videos. Additionally, following the approach in Follow-Your-Click ~\cite{ma2024followyourclick}, we calculated the Mean Squared Error (MSE) between the first frame of the generated video and the input image, with lower values indicating higher similarity. To objectively assess video loopability, we propose a metric called Loop-C, which measures the FC between the first and last frames. Higher Loop-C values correspond to better looping performance.

\subsection{Comparison Results}
We benchmarked our LoopAnimate against state-of-the-art I2V methods using objective metrics and subjective visual comparisons.
\subsubsection{Quantitative Results}
\label{sec:Quantitative Results}
We selected three open-source I2V methods, I2VGen-XL ~\cite{zhang2023i2vgenxl}, PIA~\cite{zhang2023pia}, and SVD-XT ~\cite{blattmann2023stable}, as our baselines for comparison. We conducted experiments on a test set consisting of 99 images, and the results are presented in Table~\ref{tab:Quantitative}. In Stage 3, only the \textit{conv in}, \textit{TEMM.Q}, and \textit{TEMM.V} modules were fine-tuned, achieved the best results across all metrics except Loop-C, while still being able to generate 35 video frames at once. The superior performance in CLIP-I and ${MSE}_{F_0}$ metrics indicates that our ~\textbf{LoopAnimate} generates videos with high object fidelity, supporting the second point in our contributions regarding the effectiveness of the multi-level image appearance and textual semantic decoupling framework. Furthermore, the three-stage fine-tuning strategy successfully extends the generation length to 35 frames while maintaining video quality, validating the third point in our contributions.


\begin{figure*}[t]
    \centering
    \includegraphics[width=\linewidth]{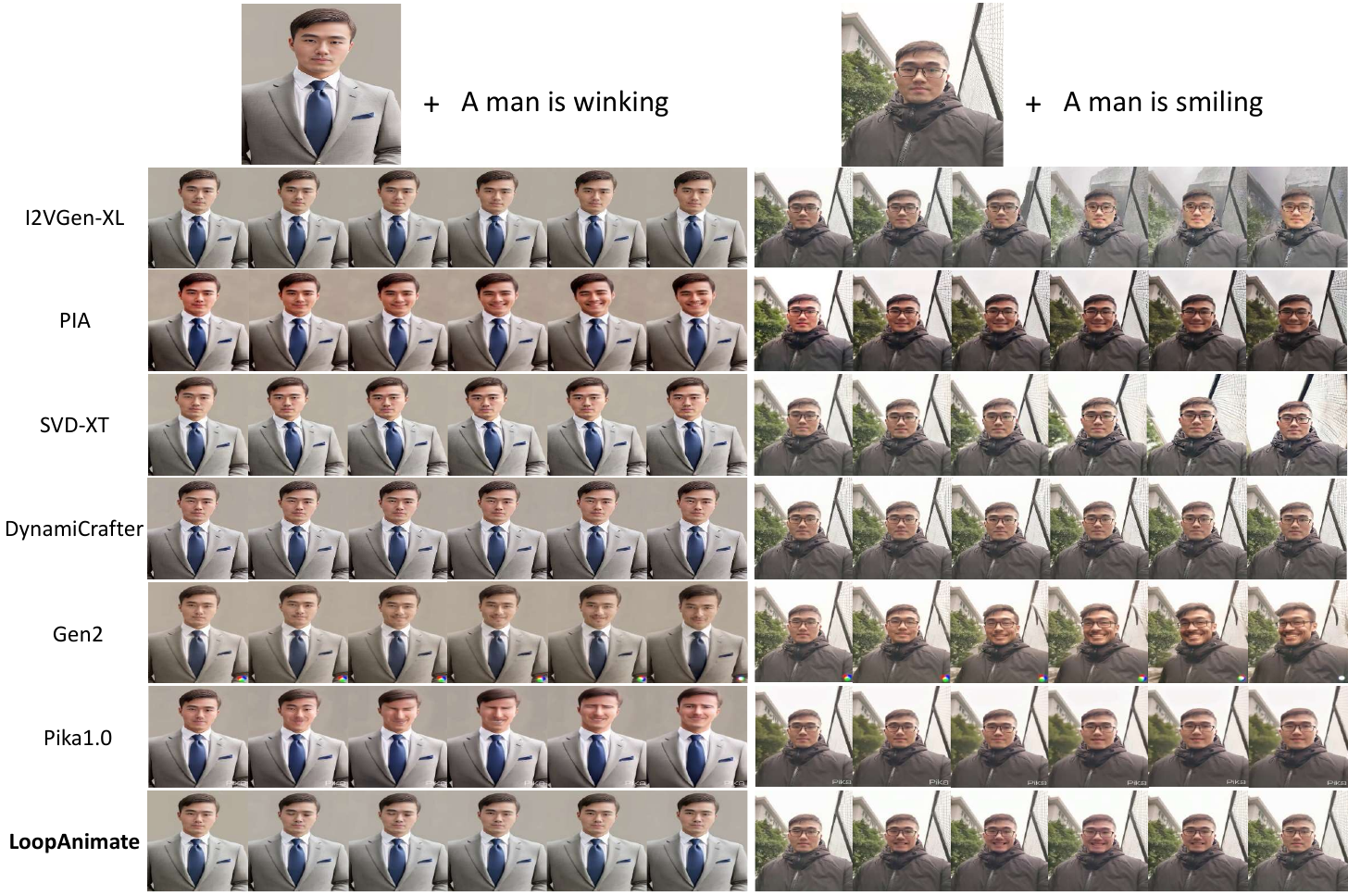}
    \caption{Visualization comparison with sota open source and commercial methods.}
    \label{fig:Fig_visualization_conparison}
\end{figure*}

\begin{figure*}[t]
    \centering
    \includegraphics[width=0.91\linewidth]{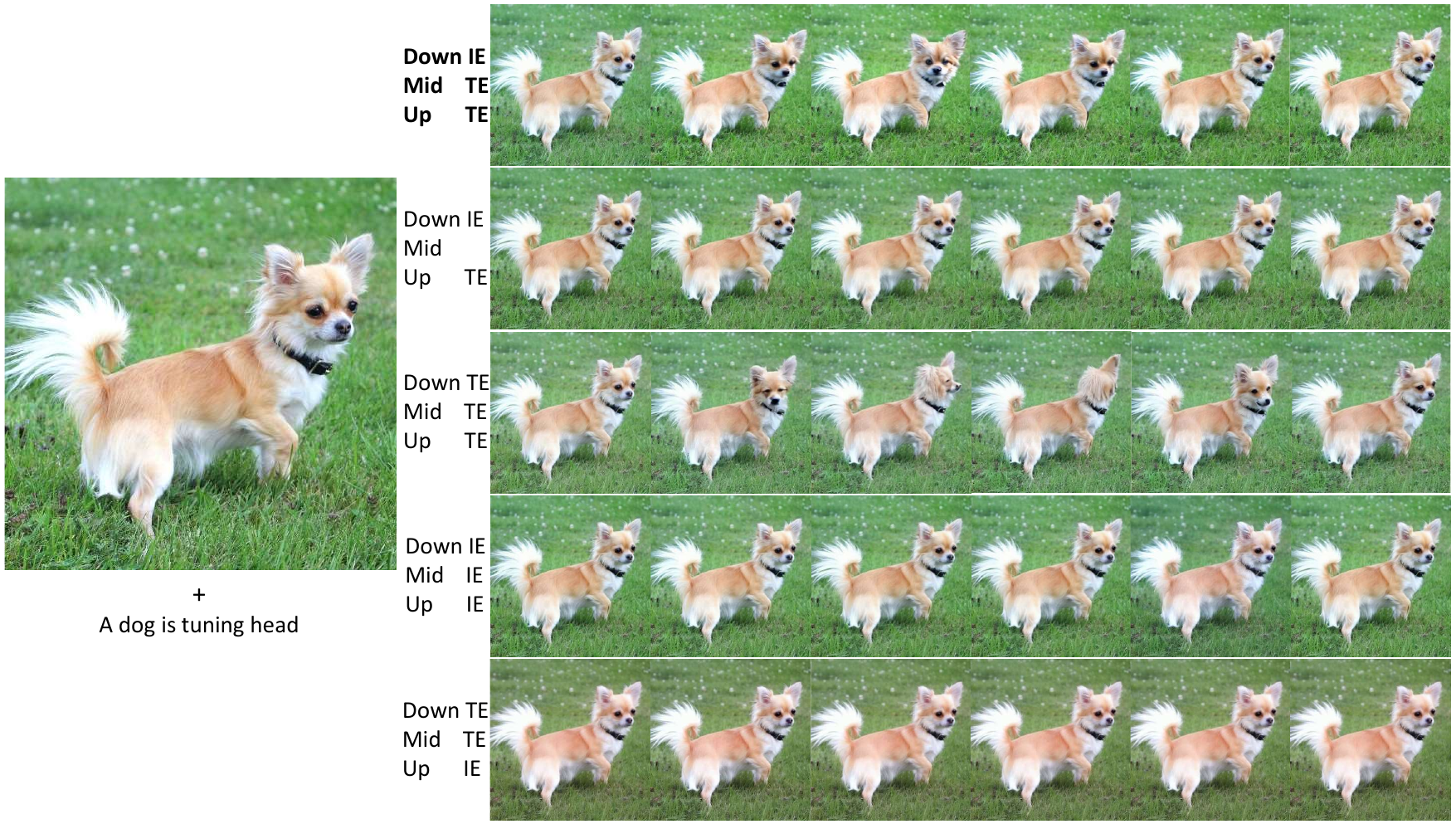}
    \caption{Visualization results of different inject position of image and text embedding. Down/Mid/Up represents down-sample/middle/up-sample block respectively. IE and TE represents image embedding and text embedding respectively.}
    \label{fig:Fig_structure_ablation}
\end{figure*}

\begin{table}[t]
\centering\small
  \setlength{\tabcolsep}{0pt} 
  \begin{tabular*}{0.95\columnwidth}{@{\extracolsep{\fill}}lccccccc@{}}
    \toprule
    \textbf{Module} &  Conv in  & CAB & TEMM & TEMM.Q &  TEMM.K &  TEMM.V\\
    \midrule
    Param(fp32)/M &  0.10  & 969.63  &1668.55 &  151.55 &  151.55 &  151.55\\
    \bottomrule
  \end{tabular*}
 \caption{Parameters of finetuning modules. }
\label{tab:parameters}
\end{table}

\begin{table}[!t]
\vspace{-2.0em}
\centering\small
  \setlength{\tabcolsep}{0pt} 
  \begin{tabular*}{0.95\columnwidth}{@{\extracolsep{\fill}}lccccccc@{}}
    \toprule
    \textbf{$Method$}   & $CLIP-I$$\uparrow$ & ${MSE}_{F_0}$$\downarrow$ & $FC$$\uparrow$ &  $Motion$$\uparrow$ & $Loop-C$$\uparrow$  & $Frames$$\uparrow$\\
    \midrule
    I2VGen-XL~\cite{zhang2023i2vgenxl} &  0.882  &54.0  & 0.972  &  0.740 &   0.812 &32 \\
    PIA~\cite{zhang2023pia}  &  0.905 &62.7  &  0.985 &  0.835 &    0.893 &16 \\
    SVD-XT~\cite{blattmann2023stable} &  0.946 &49.3 & 0.985  & 0.660  &  0.945 &25\\
    Stage1(\textbf{ours}) &  0.948 &60.6   &  0.983 & 0.781  &0.970 &15  \\
    Stage2(\textbf{ours}) &  0.953 &49.9  &  0.986 & 0.839  &  \textbf{0.971} &21 \\
    Stage3(\textbf{ours}) &  \textbf{0.956}  &  \textbf{47.2} &  \textbf{0.987} & \textbf{0.851}  &  0.969 &\textbf{35}  \\
    \bottomrule
  \end{tabular*}
 \caption{Quantitative comparison. The best average performance is in bold. ↑ indicates higher metric value and represents better performance and vice versa.}
\label{tab:Quantitative}
\end{table}

\begin{table}[!t]
\vspace{-2.0em}
\centering\small
  \setlength{\tabcolsep}{0pt} 
  \begin{tabular*}{0.95\columnwidth}{@{\extracolsep{\fill}}lcccccccc@{}}
    \toprule
    \textbf{$Method$}   & I2VGen-XL & PIA & SVD-XT &Dynamicrafter & Gen2 & Pika1.0 &  \textbf{ours} \\
    \midrule
    ${Q}_{S}$$\uparrow$ &  2.3  &1.7  & 2.7 & 2.5  &1.9 & 2.1   &\textbf{2.8}  \\
    ${Q}_{M}$$\uparrow$  & 1.8 &2.4 &  1.1 & 1.8 &1.9  &  2.3   &\textbf{2.6}   \\
    \bottomrule
  \end{tabular*}
 \caption{User study. The best average performance is in bold. ↑ indicates higher metric value and represents better performance and vice versa.}
\label{tab:User study}
\vspace{-2.5em}
\end{table}

\subsubsection{Visualization Results}
Objective metrics fail to accurately assess the subjective visual quality of generated videos. Therefore, we opted for a subjective visual comparison experiment, employing several currently available methods: the open-source I2VGen-XL~\cite{zhang2023i2vgenxl}, PIA~\cite{zhang2023pia}, SVD-XT~\cite{blattmann2023stable} and Dynamicrafter~\cite{xing2023dynamicrafter}, alongside the commercially available Runway Gen2 ~\cite{gen2} and Pika1.0 ~\cite{pikalab}. The comparison results, as illustrated in Figure.~\ref{fig:Fig_visualization_conparison}, demonstrate that our proposed LoopAnimate method exhibits superior performance in both object fidelity and motion quality, particularly in the challenging task of generating real human portraits.

Furthermore, we conducted a user study involving 19 participants. Considering the extensive task volume, we selected 20 videos from a pool of 99 (as mentioned in Sec. ~\ref{sec:Quantitative Results}) and evaluated the subjective quality based on two aspects: subject consistency ${Q}_{S}$ and motion quality ${Q}_{M}$. The scores ranged from 1 to 3, with 3 being the highest. The statistical results are presented in Table~\ref{tab:User study}. Our LoopAnimate method achieved the highest scores for both ${Q}_{S}$ and ${Q}_{M}$. It should be noted that SVD-XT ~\cite{blattmann2023stable} lacks text input. Consequently, its ${Q}_{M}$ score is significantly lower due to the evaluation's consideration of consistency with the provided textual prompts during the subjective assessment.

\begin{table}[!t]
\centering\small
  \setlength{\tabcolsep}{0pt} 
  \begin{tabular*}{0.95\columnwidth}{@{\extracolsep{\fill}}lcccccc@{}}
    \toprule
    $Index$ & $Down$ & $Middle$ & $Up$  & ${Q}_{S}$ $\uparrow$  & ${Q}_{M}$ $\uparrow$ & $Loop-C$ $\uparrow$\\
    \midrule
    \textbf{0} &  IE& TE& TE& 2.4  & \textbf{2.2}  &\textbf{0.969} \\
    1 &  IE& - & TE& 2.3  &  1.3 &0.947\\
    2 &  TE& TE& TE& 1.8  &  1.6   &0.949\\
    3 &  IE& IE& IE& \textbf{2.6}  &  1.1  &0.965\\
    4 &  TE& TE& IE& 2.1  &  1.9  &0.956\\
    \bottomrule
  \end{tabular*}
 \caption{Ablation study of inject position of image embedding(IE) and text embedding(TE). Our method is Index.0, the best average performance is in bold. ↑ indicates higher metric value and represents better performance and vice versa.}
\label{tab:stage3_ablation}
\vspace{-2.0em}
\end{table}

\subsection{Ablation study}
The ablation studies are conducted in two parts. First, we investigate the effects of incorporating image and text embeddings at different positions within the cross-attention. Second, we evaluate the impact of fine-tuning different modules during the third stage of training.

\subsubsection{Inject position of image and text embedding}
\label{sec:4.2.1}
In Sec.~\ref{sec:3.4}, we introduce the Multi-level Image representation and Textual semantics Decoupling Framework, where image and text embeddings are injected as cross-attention at different position of the denoising network. We presents an ablation study on the specific injection positions of both embeddings in Sec.~\ref{sec:4.2.1}.

The locations of image and text embedding injection within LoopAnimation are detailed in Figure.~\ref{fig:Fig_flowchat}, and determined through a combination of subjective and objective evaluations on a test set of 99 images. Subjective evaluations involved seven participants who scored the generated animations from low to high (1 being the lowest and 3 being the highest) based on two aspects: subject consistency (${Q}_{S}$) and motion quality (${Q}_{M}$). We also calculated the Loop-C objective metric to assess loopability. The results are presented in Table~\ref{tab:stage3_ablation}. The experiment group achieving the highest ${Q}_{S}$ score utilized only image embeddings. However, observations revealed minimal motion dynamics, leading to a low ${Q}_{M}$ score and a high Loop-C metric due to the restricted motion range.

The subjective visual results of the experiments are presented in Figure.~\ref{fig:Fig_structure_ablation}. Figure.~\ref{fig:Fig_structure_ablation} showcases five sets of experimental outcomes, each corresponding to the configurations detailed in Table~\ref{tab:stage3_ablation}. The structure employed within LoopAnimate is highlighted in bold within Figure.~\ref{fig:Fig_structure_ablation}. In the second row of experiments, the absence of text embedding injection into the middle block layer results in minimal motion dynamics. Additionally, the generated videos lack specific movements guided by the text prompts, which is a prevalent observation within this experimental group. This aligns with the findings of ~\cite{basu2023localizing} regarding the critical role of cross-attention within the middle block for motion generation in text-to-image tasks. Therefore, we conclude that incorporating text embedding within the middle block is equally crucial for video generation tasks.
The third row presents experiments where only text embedding information is injected into all layers. While motion dynamics are present, the lack of input image feature-level information leads to object disintegration and low-quality video generation. Conversely, the fourth row showcases experiments with exclusive image information injection into all layers, resulting in a near-complete loss of motion dynamics. This phenomenon is consistently observed in other unreported results from this group, leading to the conclusion that excessive image embedding injection can negatively impact the motion dynamics of the generated output.

\begin{table}[!t]
\centering\small
  \setlength{\tabcolsep}{0pt} 
  \begin{tabular*}{0.95\columnwidth}{@{\extracolsep{\fill}}lcccccc@{}}
    \toprule
    \textbf{$TEMM$} & $CLIP-I$ $\uparrow$& ${MSE}_{F_0}$$\downarrow$  & $FC$ $\uparrow$ &  $Motion$ $\uparrow$ & $Loop-C$ $\uparrow$  & $Frames$$\uparrow$\\
    \midrule
    Q &  0.937  &52.3  &  0.978 & 0.781  &  0.968  &35\\
    QK &  0.943 &53.5  &  0.979 & 0.811  &  0.968   &35\\
    \textbf{QV} &  0.956 &\textbf{47.2}   &  \textbf{0.987} & 0.851  &  \textbf{0.969}  &35 \\
    QKV &  \textbf{0.958} &51.1   &  \textbf{0.987} & \textbf{0.873}  &  0.968 &35 \\
    \bottomrule
  \end{tabular*}
 \caption{Ablation study of different finetuning modules in TEMM. The best result is in bold. ↑ indicates higher metric value and represents better performance and vice versa.}
\label{tab:structure}
\vspace{-3em}
\end{table}

Finally, the last row displays experiments with text embedding injection into the down-sample and middle layers, while the up-sample layer receives image embedding. This configuration yields suboptimal motion dynamics and image quality. The primary issue observed within the remaining experiments of this group is the unsatisfactory image quality.
Based on the comprehensive analysis of statistical results derived from $99*5$ videos across the five sets of experiments described above, the structure of LoopAnimate is determined as elaborated in Sec.~\ref{sec:3.4}.

\subsubsection{Finetuning modules in TEMM}
\label{sec:4.2.2}
Following the analysis in Sec.~\ref{sec:3.5}, we hypothesize that fine-tuning only the TEMM.Q and TEMM.V components during the third stage is sufficient to achieve the desired results. Sec.~\ref{sec:4.2.2} presents experimental validation of this hypothesis, with results summarized in Table~\ref{tab:structure}.

The results demonstrate that fine-tuning either $QV$ or $QKV$ within \textbf{TEMM} yields comparable performance in terms of objective metrics. Specifically, the object fidelity metric (CLIP-I), frame coherence metric (FC), and video loopability metric (Loop-C) exhibit minimal discrepancies between the two approaches. However, fine-tuning $QKV$ leads to superior results for the motion quality metric, while fine-tuning $QV$ excels in terms of the similarity between the first video frame and the input image. Notably, all experiments achieve high Loop-C scores, validating the first contribution stated in the introduction: the capability to generate loopable videos.

\section{Discussion and conclusion}
This paper proposes LoopAnimate, a novel video generation method capable of producing videos with seamless loops between the first and last frames. Notably, LoopAnimate extends the one-shot video generation length to 35 frames, which is a significant advancement in the field. We conducted extensive experiments on a in-house dataset of subject images, comparing LoopAnimate with existing image-to-video methods. Both objective and subjective evaluation results demonstrate that LoopAnimate outperforms current approaches in terms of subject preservation, inter-frame continuity, and motion quality, validating its effectiveness.


\bibliographystyle{IEEEtran}


\end{document}